\documentclass[]{interact}

\usepackage{epstopdf}
\usepackage[caption=false]{subfig}

\usepackage[natbibapa,nodoi]{apacite}
\theoremstyle{plain}
\newtheorem{theorem}{Theorem}[section]

\newtheorem{hypothesis}[theorem]{Hypothesis}
\newtheorem{fact}[theorem]{Fact}
\newtheorem{remark}[theorem]{Remark}

\theoremstyle{definition}

\begin{document}


\title{Arguments for the Effectiveness of Human Problem Solving}

\author{
\name{Frantisek Duris\thanks{Email: fduris@dcs.fmph.uniba.sk}}
\affil{Faculty of Mathematics, Physics and Informatics,\\ Comenius University, Bratislava, Slovakia}
}

\maketitle

\begin{abstract}
The question of how humans solve problem has been addressed extensively. However, the direct study of the effectiveness of this process seems to be overlooked. In this paper, we address the issue of the effectiveness of human problem solving: we analyze where this effectiveness comes from and what cognitive mechanisms or heuristics are involved. Our results are based on the optimal probabilistic problem solving strategy that appeared in Solomonoff paper on general problem solving system. We provide arguments that a certain set of cognitive mechanisms or heuristics drive human problem solving in the similar manner as the optimal Solomonoff strategy. The results presented in this paper can serve both cognitive psychology in better understanding of human problem solving processes as well as artificial intelligence in designing more human-like agents.
\end{abstract}


\section{Introduction}

Problem solving is probably the most common activity of all organisms, especially of humans. We deal with various problems throughout our lives, from infancy to adulthood. Therefore, it is not surprising that an extensive effort has been made to understand the cognitive processes responsible for this ability, and many models of human problem solving have been presented \citep{pol2, ste2, ohl, wei, ste, hum, woods, ste3}. Additional models of problem solving and, generally, of human cognitive skills have been proposed by research groups on cognitive architectures like ICARUS \citep{lan, lan5, lan3, lan2}, ACT-R \citep{and2, and}, SOAR \citep{nas, lai, lan3}, Polyscheme \citep{cas, cas2}, and others (see \citealp{duch, lan3}). However, while the current state of research can explain some questions about the scope of problems the human brain can solve, we lack some form of explanations what makes solving (novel) problems so fast, i.e., effective.  We note that there have been proposed models or algorithms for general problem solving with special focus on the provable effectiveness \citep{sol, hutter2002fastest, oops, fink,marcus2005universal}, but these models are not designed to mimic human problem solving process, and as such offer little explanation of this feat. Thus, the question why humans are still more effective problems solvers than computers is still left open \citep{poole2010artificial}. 

In this paper, we present some ideas on this problem. The core of our arguments revolves around an optimal betting strategy for the following scenario. You are in the gambling house making bets. All bets win the same prize but they have different probability $p_i$ of winning and come at different cost $d_i$. If you can make each bet only once, what strategy will give you the greatest win probability per dollar? According to \citet{sol}, the best strategy is to sort the bets by the ratio $p_i/d_i$ and take them one by one starting from the top. We argue how this strategy relates to the human problem solving process, and what does it mean for its effectiveness. 

The paper is organized as follows. First, we outline what is an optimal problem solving strategy in the sense of Solomonoff betting strategy. In the subsequent six sections, we argue how particular six cognitive mechanisms play role in driving human problem solving by the same manner. Finally, we close the paper with some remarks on the applicability of those six cognitive mechanisms in problem solving.

\section{The effective problem solving strategy} \label{sub:basic}

In the following text we will use the term \emph{solution candidate} as any sequence of steps the solver can consciously take during the problem solving process. Examples being a particular rotation of Rubik cube, Google search for an existing solution, or drawing a visual representation of a problem and then rubbing out some parts of it. Intuitively, the notion of solution candidate represents a process, method, or idea how to proceed in problem solving, which also includes how the solver generates (or derives, or constructs) and selects the information used in this process. Thus, we have that the same process, method, or idea coupled with different approach to generation or selection of the used information represent different solution candidates. Moreover, we do not require the solver to know how to perform these steps (that can be a different problem on its own), but they should be clear enough to articulate or at least to understand. One can observe that in machine problem solving, a solution candidate is an analogue to a formal string of characters that can be fed to a Turing machine to check as a problem solution. 

We realize that this definition of the solution candidates is still a little bit vague but this is probably as concrete as it gets in cognitive psychology. Our defense is that too much formalism can lead to deviation from the actual human problem solving towards machine problem solving. In this sense, let us call by \emph{Blind search} a method of solving problems by checking all potential solution candidates in no particular order. By potential solution candidates we mean all the solution candidates the solver can think of at some point in the problem solving process. In machine problem solving, this method can solve any problem (provided the problem has a finite solution), although the time required can be huge. The same holds for human problem solving where the solver can, in principle, write down each combination of words, and check if they put together a valid solution. On the other hand, the following Solomonoff strategy presents a probabilistically optimal, i.e., effective, approach with a high chance of finding a solution in a reasonably short time.

\begin{theorem}[\citealp{sol}]
Let $m_1, m_2, m_3, ...$ be candidates that can be used to solve a problem. Let $p_k$ be probability that the candidate $m_k$ will solve the problem, and $t_k$ the time required to test this candidate. Then, testing the candidates in the decreasing order $\frac{p_k}{t_k}$ gives the minimal expected time before a solution is found.
\label{thm:sol}
\end{theorem}

\begin{remark}
To set things right, this strategy was probably known before Solomonoff used it in his general problem solving system. However, as he did not mention any references regarding this strategy (Theorem I) in his paper \citep{sol}, or, for that matter, a proof, which can now be found in \citep{duris2016error}, and we were not able to trace this strategy to any other source, we settled for the term Solomonoff (optimal) problem solving strategy.
\end{remark}

It follows that the effectiveness of problem solving depends on 
\begin{enumerate}
	\item the database of solution candidates, i.e., knowledge and experience of the solver,
	\item the ability to generate in the effective order (\emph{Theorem \ref{thm:sol}}) and in a short time the appropriate solution candidates.
\end{enumerate}
Thus, there are differences between human solvers based on their knowledge \citep{voss1983problem} as well as their cognitive skills \citep{bas}. For this reason, we will talk about the problem solving effectiveness from the solver's perspective rather than the objective effectiveness of the found solution. Therefore, both novice and expert can solve a given problem effectively even though the objective effectiveness of their solutions might be (vastly) different.

The problem solving process is a recursive one, i.e., solving one problem can create new (sub)problems which have their own sets of solution candidates, and the probabilistically fastest way to solve them is again by Solomonoff strategy. We note that \emph{Theorem \ref{thm:sol}} assumes that the solver knows the values of $p_i$ and $t_i$ exactly what is rarely the case in real life. Instead, the solver may only know approximations of these values. For example, he may only know that a particular method is not very reliable, i.e., it has low value $p_i$, or that it is very time consuming, i.e., large value $t_i$. However, this is not necessarily a problem in our theory because the solver usually has, at some point during the problem solving process, only a handful of methods (solution candidates), and this approximate knowledge about such methods can still help him to apply them in the effective order according to \emph{Theorem \ref{thm:sol}}. Moreover, in the following sections we will argue that human brain has several subconscious abilities to help the solver to generate solution candidates in the effective order even without the explicit knowledge of the values $p_i$ and $t_i$. 

However, we do not wish to argue that human problem solving is as mathematically optimal as Solomonoff strategy. It is most likely not. Rather, we wish to simply point out that there are several cognitive mechanisms that, on average, seem to drive human problem solving process in the direction similar to that of \emph{Theorem \ref{thm:sol}}, and, because of this, they are most likely (part of) the source of the effectiveness of human problem solving. For this reason, these mechanisms should be considered for implementation in cognitive architectures aiming at human-like intelligent agents.

\begin{hypothesis}
The following mechanisms play a key part in the effectiveness of human problem solving:
\begin{enumerate}
	\item {Discovering similarities,}
	\item {Discovering relations or associations,}
	\item {Generalization,}
	\item {Abstraction,}
	\item {Intuition,}
	\item {Context sensitivity.}
\end{enumerate}	
\label{prop:effmech}
\end{hypothesis}

In the following six sections we present arguments for each item on the list that it participates in the effectiveness of human problem solving with respect to \emph{Theorem \ref{thm:sol}}.


\section{Discovering similarities} \label{sub:sim}

\begin{fact}
Similar problems often have similar solutions.
\label{fact1}
\end{fact}

A process of solving problems based on the solutions of the similar problems solved in the past is also called Case-based reasoning \citep{slade}. The \emph{Fact \ref{fact1}} is used by everyone on a daily basis to quickly solve many situations in life, work, etc. It is not hard to see that it also fits perfectly with the Solomonoff strategy. More particularly, the solver identifies a set of problem concepts (properties, features, structure etc.) based on which he can quickly access similar concepts stored in his memory that are associated with an idea how to solve a known problem. In such a case the new problem and the known one are similar via this set of concepts. According to the \emph{Fact \ref{fact1}}, this process provides the solver with highly relevant solution candidates (i.e., with relatively high values  $p_i/t_i$), and, vice versa, the solution candidates that are not similar with the current problem are not recalled in this way. Moreover, it was observed that humans have cognitive abilities that support effective search for similarities and patterns \citep{bejar, gent, rob}. Thus, this process is relatively quick and often successful, which is in accord with (ii) above.


\section{Discovering relations or associations} \label{sub:rca}

\begin{fact}
Related facts, problems or situations (i.e., relevant information) often hold the clues for the solution of the problem.
\label{fact2}
\end{fact}

A problem involving a right triangle can fire up an association with the Pythagorean theorem which can be used in search of a solution (the right triangle and the Pythagorean theorem are closely related but not really similar). \citet{alt} developed TRIZ, a general strategy for creative problem solving, that exploits already solved problems to suggest new solution candidates, and other authors also rely on finding related problems and situations in their problem solving strategies \citep{pol2,woods}. 

Clearly, related facts, problems, situations, experiences etc. (i.e., relevant information) often hold crucial clues for the solution of the current problem without which the solver would get stuck. Interestingly, human brain seems to be wired in a way not only to store concepts but also relationships among them. Thus, it is able to support quick retrieval of highly relevant information (not necessarily similar as in \emph{Section \ref{sub:sim}}). Evolutionary benefits of such case are obvious, and there is a solid theoretical support for it. A semantic network is a well established model of the long term memory \citep{qui,collins1969retrieval}. Concepts, representing knowledge and experience are joined by associations of potentially different weight. Concepts are activated by a process of spreading activation from the already active concepts through their weighted associations \citep{col,anderson1983spreading, ohl}. The adjustments of the weights of the associations as well as their creation/deletion (forgetting) is accomplished based on the Hebb rule \citep{hebb}. It follows that by this rule we have the strongest associations among the most related concepts which again fits with the Solomonoff strategy. More precisely, the ability of human brain to quickly retrieve relevant information together with \emph{Fact \ref{fact2}} mean that the solver has relatively quick access to relevant information which often contain crucial information for the solution, and with which the solver can more easily and quickly solve the problem. Such approach to problem solving is thus often quick and successful, and by this it is in accord with (ii) above.

Observe also that even though human brain is a vast storage of various interconnected knowledge and experiences, by Hebb rule there are no strong connections between totally unrelated concepts. In this way, the human brain instantaneously filters out huge amounts of information which are most likely unimportant for the solution of the current problem. This may not be always the case, but recall that the Solomonoff strategy is probabilistic. 

\section{Generalization} \label{sub:gen}

It was observed that humans are more likely to remember only the crucial parts of information or experience rather than specific details \citep{censor2013generalization,xu2013neural}. In this sense, we are more likely to remember crucial steps in a complicated construction like a mathematical proof rather than the exact details. On one hand, it is inefficient to store too much information that can be easily filled in on-demand. On the other hand, proofs reduced to a few key steps are more likely to fit other problems as well. For example, dynamic programming originated as a solution to one particular problem but in time it was generalized to a paradigm applicable on a whole range of different problems. Thus, generalization helps human solvers to overcome the threshold of similarity (between two problems, or, in general, two concepts) by reducing the number of parts that needs to fit which in turn helps the effectiveness of problem solving by the same reason as discovering similarities.

Furthermore, we can imagine problem solving as browsing an association tree where we have information associated with the identified problem properties. By browsing such tree, we are trying to find some crucial information for the solution. In case of un\-ge\-ne\-ra\-li\-zed specific past problems, solutions, proofs etc. we would browse large trees full of unimportant details which can take a long time. By generalization, we can prune these trees to much smaller thus making problem solving faster, i.e., more effective. Of course, we may not find the right information always, but by Solomonoff strategy it pays, on average, to check the most promising information (i.e., candidates) first.

\section{Abstraction} \label{sub:abs}

\begin{fact}
Problems with the same gist/structure often have the same or a similar solution.
\label{fac4}
\end{fact}

Abstraction is extracting from a given problem representation only those features that are related to the essence of the problem, i.e., its general pattern \citep{adler1984abstraction}. For example, computing the life expectancy of a fly that is zig-zagging between two trains going in the opposite direction is in fact a simple ``distance = speed x time'' problem, if we abstract from the shape of the fly's trajectory. In this sense, working with abstractions reduces the problem space as many problems coalesce into one abstract version. On one hand, using the reduced (i.e., abstract) problem space allows more effective comparisons between the problems. On the other hand, it helps human solvers to further overcome the threshold of similarity (between two problems, or, in general, two concepts) because superficially different problems can be classified as similar on a deeper level. Thus, abstraction connects the human problem solving effectiveness with the effective Solomonoff strategy, and improves the problem solving effectiveness by the same way as does discovering similarity and generalization described in \emph{Sections \ref{sub:sim}} and \emph{\ref{sub:gen}} above.

Moreover, abstraction enables the solver to direct his focus on the essential features of the problem instead of surface details. In this way, the solver does not waste time by exploring irrelevant (e.g., surface) problem features or associations, i.e., weak solution candidates. As was the case with generalization, this may not always work, i.e., it is a bet. But, according to the \emph{Fact \ref{fac4}}, it is a bet with a good chance of success.

\section{Intuition} \label{sub:int}

The role of intuition in problem solving and decision making is undeniable \citep{style1979intuition,metcalfe1987intuition,dorfman1996intuition,pretz2008intuition,lufityanto2016measuring}. However, we do not mean to discuss how does it work here. For our purposes, intuition is something similar to a preschool child learning a grammar. The child can use the rules mostly correctly, but he or she cannot formulate them. However, the rules are in some accessible way stored in the child's memory. In this way, intuition provides the solver with candidates to test or associations to follow even though the solver cannot articulate the reasons why he should do so (i.e., such candidates may not appear to the solver visibly similar or related to the current problem). In words of \citet{nelissen2013intuition}, \emph{problem solver -- using the available, compacted knowledge -- is aware of the direction in which the thinking process should proceed and what kind of information and knowledge can be relevant.} Thus, it follows that intuition connects the human problem solving with the effective Solomonoff strategy as does discovering similarities and relations/associations described in \emph{Sections \ref{sub:sim}} and \emph{\ref{sub:rca}} above.

\section{Context sensitivity} \label{sub:cont}

Concepts in solver's memory have, in general, many associations \citep{ohl}, and each such association with the given problem is a potential solution candidate. However, those candidates that do not fit the context in which the problem is set are probably not going to be part of the solution. Imagine again the problem solving as browsing an associations tree where we have information associated with the identified problem properties. By (temporarily) blocking the associations that do not fit the context, the association tree in which the solver is searching for a crucial information is pruned. In other words, the solver's focus is directed to the most promising associations (solution candidates). This again fits with the Solomonoff strategy that candidates most likely to succeed are checked first.

\section{Conclusion}

We proposed a connection between six well-known cognitive mechanisms (discovering similarities, discovering relations or associations, generalization, abstraction, intuition, context sensitivity) and Solomonoff optimal problem solving strategy, which states that if there are multiple approaches to finding a solution to the problem, then one should start with the most promising approach first (of course, respecting application times of the approaches as well). In this paper, we put forward arguments that these six mechanisms drive human problem solving in the very similar manner, which may be the first explicit argument for the effectiveness of human problem solving. We stress again that the effectiveness is meant from the solver's perspective (i.e., given his knowledge and the state of his cognitive skills). Thus, both novice and expert can solve a given problem effectively even though the objective effectiveness of their solutions might be (vastly) different.

Furthermore, the presented mechanisms are those for which we found a direct relation with the Solomonoff strategy. However, this does not imply that these six mechanisms act alone in the process of solving problems. They are supported by cognitive facilities for language processing, working memory system for coordinating, monitoring and executing the intended actions, long-term memory system for retrieval and interpretation of memories and experiences and so on. 

Also, \citet{cas} proposed an idea that only a few cognitive mechanisms are needed to explain the domain-universal human intelligence. His initial guess what such mechanisms could be includes reasoning about time, space, parthood, categories, causation, uncertainty, belief, and desire. While they do not match our list from \emph{Hypothesis \ref{prop:effmech}} (Cassimatis talks about the domain-universal human intelligence, not the effectiveness of problem solving), they give support to the idea that effective problem solving does not need to entail overly complex models.

The presented results in this paper can serve two purposes. First, based on this preliminary work, we can start building a more robust understanding of how people solve problems effectively. In this sense, we can further research the following issues. Is the set of mechanisms in \emph{Hypothesis \ref{prop:effmech}} complete? Given some support processes (e.g., language processing, working and long-term memory processes), how large role do mechanisms from \emph{Hypothesis \ref{prop:effmech}} play in problem solving? Are there any other key mechanisms in problem solving which we did not cover? Second, proper understanding of principles behind the effectiveness of human problem solving can help the design of cognitive architectures (or AI generally) to develop agents with more human-like abilities. If some cognitive architectures already support some of these mechanisms, then this work can act as an argument about their problem solving effectiveness.

\bibliographystyle{apacite}
\bibliography{bibi}

\end{document}